\pdfoutput=1

\documentclass[11pt, dvipsnames,table]{article}

\usepackage[final]{acl}

\usepackage{times}
\usepackage{latexsym}

\usepackage[T1]{fontenc}

\usepackage[utf8]{inputenc}

\usepackage{microtype}

\usepackage{inconsolata}

\usepackage{graphicx}
\usepackage{multirow}
\usepackage{multicol} 
\usepackage{tabularx}
\usepackage{makecell}
\usepackage{enumitem} 
\usepackage{amsmath}
\usepackage{subcaption}
\usepackage{tcolorbox} 
\usepackage{graphicx}

\usepackage{booktabs}  
\usepackage{geometry}
\usepackage{amssymb}
\usepackage{hhline}  

\usepackage{algorithm}
\usepackage{algpseudocode}

\title{Zero-Shot Fine-Grained Image Classification \\ Using Large Vision-Language Models \vspace{-0.75cm}}

\author{Md. Atabuzzaman \qquad Andrew Zhang \qquad Chris Thomas \\
Department of Computer Science\\
  Virginia Tech \\ 
  \texttt{\{atabuzzaman, azhang42, christhomas\}@vt.edu}\\}

\begin{document}
\maketitle

\begin{abstract}

Large Vision-Language Models (LVLMs) have demonstrated impressive performance on vision-language reasoning tasks. However, their potential for zero-shot fine-grained image classification, a challenging task requiring precise differentiation between visually similar categories, remains underexplored. We present a novel method that transforms zero-shot fine-grained image classification into a visual question-answering framework, leveraging LVLMs' comprehensive understanding capabilities rather than relying on direct class name generation. We enhance model performance through a novel attention intervention technique. We also address a key limitation in existing datasets by developing more comprehensive and precise class description benchmarks. We validate the effectiveness of our method through extensive experimentation across multiple fine-grained image classification benchmarks. Our proposed method consistently outperforms the current state-of-the-art (SOTA) approach, demonstrating both the effectiveness of our method and the broader potential of LVLMs for zero-shot fine-grained classification tasks. Code and Datasets: \url{https://github.com/Atabuzzaman/Fine-grained-classification}

\end{abstract}

\section{Introduction}
\begin{figure}[!ht]
    \centering
    \includegraphics[width=\linewidth]{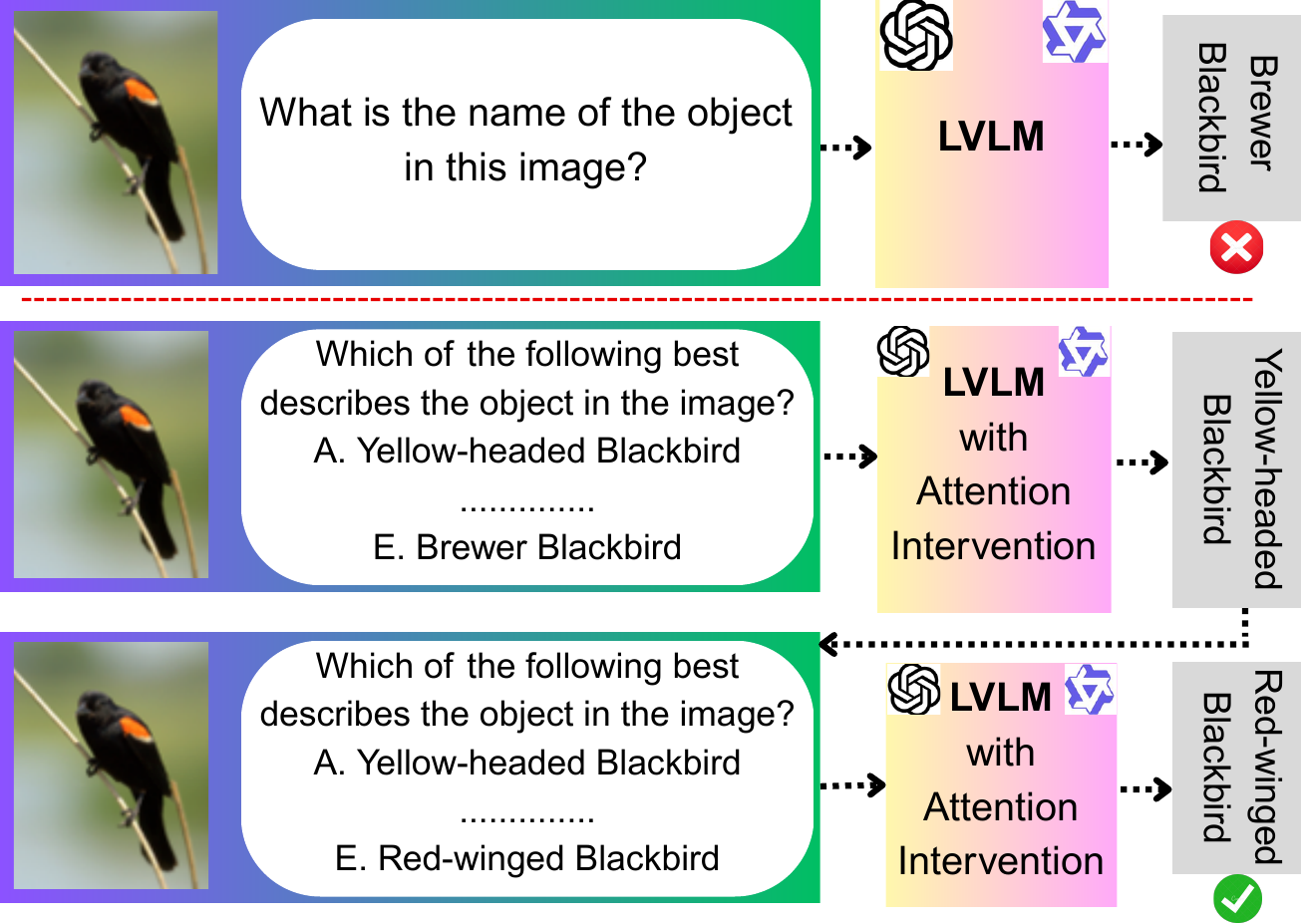}
    \caption{Overview of our zero-shot fine-grained image classification framework. Unlike existing approaches (top), which directly prompt a Large Vision-Language Model (LVLM) to generate a class name given an input image, our method leverages an LVLM combined with a proposed iterative multiple-choice question-answering strategy and an attention intervention technique to select the most accurate fine-grained class description. This framework effectively matches each input image with the most appropriate class description without requiring any training samples.}
    \label{fig:concept}
    \vspace{-0.5cm}
\end{figure}

Large Vision-Language Models (LVLMs) such as LLaVA~\cite{llava2,llava}, BLIP-2~\cite{blip2}, InstructBLIP~\cite{instructblip}, Molmo~\cite{molmo}, Qwen2-VL~\cite{qwen2}, InternVL~\cite{internvl2_5}, and DeepSeek's Janus-Pro~\cite{januspro} and so on, alongside closed-source models like GPT-4V~\cite{gpt4o} and others have revolutionized multimodal understanding. These models demonstrate remarkable zero-shot capabilities in tasks that require extensive real-world knowledge, such as Visual Question Answering (VQA), visual reasoning, and image captioning. Their success largely stems from instruction-tuning and pre-training on vast multimodal datasets, enabling effective integration of visual content with textual knowledge.

\begin{figure}[!ht]
    \centering
    \includegraphics[width=1.0\linewidth]{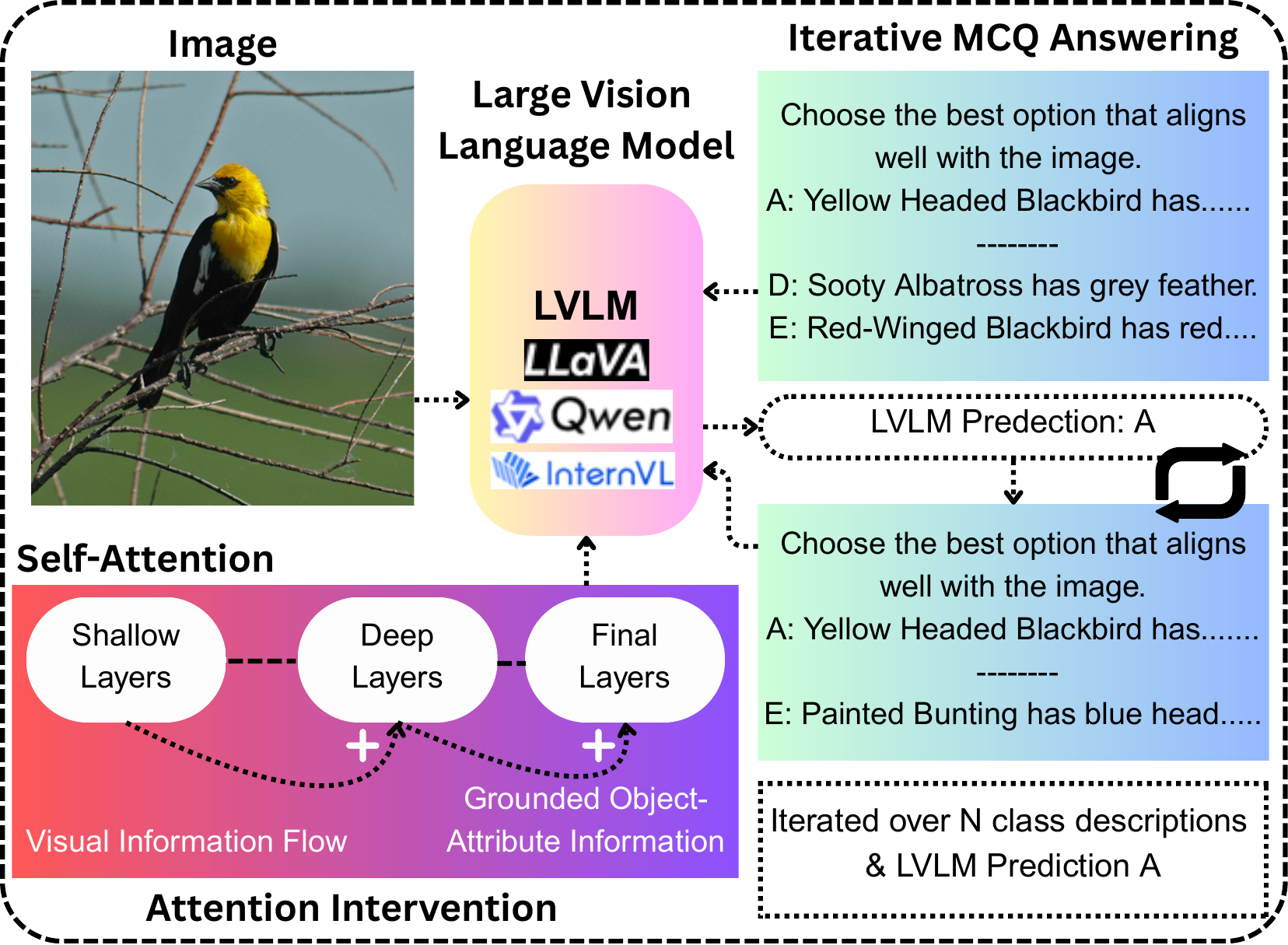}
    \caption{Overview of our proposed zero-shot fine-grained image classification framework using LVLM. The system takes an input image and class descriptions with a prompt, and uses an LVLM enhanced with an attention intervention mechanism. The framework employs an iterative MCQA approach where the LVLM selects the most appropriate class description through multiple rounds of refinement. The attention intervention module guides the visual information flow from shallow to deep layers, while deep layers provide grounded object-attribute information to final layers to improve classification accuracy.}
    \label{fig:overview}
    \vspace{-0.5cm}
\end{figure}

However, despite their impressive performance in general vision-language tasks, LVLMs face significant challenges in fine-grained image classification, which requires distinguishing subtle visual differences between highly similar categories, such as bird species or car models~\cite{finer}. Unlike coarse-grained tasks where broad visual features suffice, fine-grained classification demands a precise understanding of nuanced visual characteristics and their alignment with detailed textual descriptions (Figure~\ref{fig:concept} \&~\ref{fig:overview}). This challenge becomes particularly acute in zero-shot settings, where models must identify fine-grained categories without task-specific training.

Initial attempts to address this challenge include Finer~\cite{finer}, which trained models to generate fine-grained attributes before predicting class names, and GRAIN~\cite{grain}, which adapted CLIP~\cite{clip} with LVLMs to measure similarity between generated text and ground truth class names. While these methods represent important first steps, their limited performance reveals critical gaps in current LVLM architectures and methodological approaches.
Although LVLMs excel at generating descriptive captions and answering general questions about images, they struggle to capture and differentiate the subtle visual distinctions that define fine-grained categories~\cite{finer}. This limitation stems from their reliance on direct class name generation without sufficient domain knowledge and the use of less powerful models. These findings suggest that vision-language generation approaches may not adequately address the level of detail required for fine-grained classification tasks.

To address these limitations, we propose a novel method for zero-shot fine-grained image classification using LVLMs. Inspired by VQAScore~\cite{cmu}, we first develop a strong foundation through likelihood-based yes/no QA classification. We then introduce an iterative multiple-choice question-answering (MCQA) approach to enable fine-grained classification. To reduce the computational overhead of iterative MCQA, we also explore an all-at-once approach that leverages the extended context capabilities of modern LVLMs to process all class descriptions simultaneously in a single forward pass. Rather than relying on direct class name generation, both approaches leverage the models' comprehensive visual understanding capabilities. Our method aligns with traditional CLIP-based fine-grained classification approaches~\cite{menon2022visual, saha2024improved} by utilizing class labels and descriptions. Additionally, inspired by recent advances in LVLM hallucination reduction~\cite{pai, causalmm, opera, deCo, jiang2024interpreting}, we introduce a simple yet effective attention intervention technique. This intervention integrates insights from multiple studies, combining early-layer visual information flow~\cite{zhang2024redundancy} with deep-layer ground truth token detection capabilities~\cite{deCo, jiang2024interpreting} to enhance classification accuracy.

Furthermore, we identify a critical limitation in existing class descriptions, which are typically generated using Large Language Models (LLMs) such as GPT-3/4~\cite{menon2022visual, saha2024improved} or sourced from the internet and refined using GPT-4V~\cite{finer}. These descriptions often contain visual attributes that are absent from the actual class images or include irrelevant information that may impair model performance. To address this challenge, we develop curated class description benchmarks that more accurately capture the visual characteristics present in the images. Our approach investigates settings both with and without explicit class names in the descriptions, providing greater flexibility and robustness in classification tasks. Through extensive experiments on major fine-grained classification benchmark datasets, our methods demonstrate significant performance improvements over existing approaches and achieve SOTA performance. The main contributions of our work are as follows:
\begin{itemize}[noitemsep, topsep=0pt]
    \item We introduce a novel method for zero-shot fine-grained image classification that leverages LVLMs' understanding capabilities rather than relying on class name generation.
    \item We propose an efficient attention intervention technique that enhances visual information flow by combining early-layer visual information with deep-layer semantic understanding.
    \item We introduce improved class description benchmarks that better align with visual features, addressing gaps in existing descriptions.
    \item We demonstrate substantial performance improvements over existing methods across five major fine-grained classification benchmarks.
\end{itemize}


\section{Related Work}

\subsection{Zero-shot Image Classification using Vision-Language Models}
Vision-Language Models (VLMs) are trained on massive web-scraped datasets for zero-shot image classification. CLIP~\cite{clip} pioneered this approach through contrastive training of image and text encoders, optimizing for high similarity between matching pairs while minimizing similarity for non-matching ones. During inference, CLIP computes similarities between unseen images and text captions to perform classification. ALIGN~\cite{align} extended this framework by adding a data refinement stage.

Recent approaches leverage LLMs to generate richer class descriptions. \citet{menon2022visual} prompted LLMs to generate discriminative attributes for each class, while \citet{pratt2023does} proposed generating multiple prompts per class to enrich text understanding. \citet{saha2024improved} fine-tuned CLIP using attribute-based descriptions incorporating visual features, geographic information, and habitat context. \citet{chiquier2024evolving} enhanced interpretability by combining concept bottlenecks~\cite{koh2020concept} with evolutionary search for LLM-generated concepts. These methods typically determine the predicted class through maximum cosine similarity between CLIP-generated text and image embeddings.

\subsection{Zero-shot Image Classification using LVLMs}

Recent advancements in LVLMs including LLaVA~\cite{llava2,llava}, BLIP-2~\cite{blip2}, InstructBLIP~\cite{instructblip}, and Qwen2-VL~\cite{qwen2} have depicted remarkable zero-shot capabilities in tasks like Visual Question Answering (VQA), visual reasoning, and image captioning. However, their ability to classify fine-grained images remains underexplored.

Prior work in zero-shot fine-grained classification with LVLMs includes Finer~\cite{finer}, which generates class names and considers a prediction correct if the ground truth appears among the top 20 generated tokens. GRAIN~\cite{grain} extended this approach by computing cosine similarity between CLIP embeddings of LLaVA-v1.6 generated text~\cite{llava} and ground truth class descriptions. In contrast, following traditional zero-shot classification paradigms, our work introduces a novel method that leverages class descriptions instead of generating class names. We provide these descriptions along with the image as input to LVLMs. We either measure the probability of a `Yes' response based on image-text alignment or prompt the model to select the most image-aligned description.

\subsection{Attention Intervention}

Recent research has explored various techniques to mitigate LVLM hallucinations in training-free settings. These approaches include visual contrastive decoding (VCD)~\cite{vcd}, token selections~\cite{sid}, overconfident token penalization using attention weights~\cite{opera, damro}, and attention intervention~\cite{pai,causalmm}. PAI~\cite{pai} demonstrated that LVLMs tend to underweight image tokens, with language decoders showing bias toward text tokens. To address this, PAI increased image token attention while subtracting language bias, following VCD~\cite{vcd}. CAUSALMM~\cite{causalmm} employed various attention intervention strategies (random, uniform, reversed, and shift) and counterfactual reasoning at both visual and language attention levels to mitigate modality priors and enhance input-output alignment. DeCo~\cite{deCo} proposed dynamic selection of previous layers to incorporate knowledge into final layers, while \citet{jiang2024interpreting} introduced a system for hallucination mitigation through interpretation and editing of image embeddings. However, these methods are computationally expensive~\cite{deCo, jiang2024interpreting}, as they rely on visual contrastive decoding and token or layer selection. This inefficiency makes these methods unsuitable for our classification task, which involves comparing a single image against multiple class descriptions. Instead, our method achieves greater efficiency by using a single forward pass while propagating visual information from shallow to deep layers and visual object-attribute grounded information (via attention weights) from deep layers to final layers. Our approach significantly reduces computational overhead while maintaining effectiveness in zero-shot fine-grained image classification.
\vspace{-0.1cm}

\section{Proposed Method}

This section presents our proposed method for zero-shot fine-grained image classification using LVLMs. We begin by introducing a likelihood-based Yes/No Question-Answering (QA) approach, then propose an iterative MCQA framework to address the limitations of the initial approach. We introduce a simple yet effective attention intervention technique inspired by hallucination mitigation methods to enhance iterative MCQA approach performance. Figure~\ref{fig:overview} provides an overview of our iterative MCQA method with attention intervention. Finally, we discuss an all-at-once approach that leverages the extended context capabilities of modern LVLMs to reduce computational overhead.


\subsection{Yes/No Question-Answering}

Inspired by VQAScore~\cite{cmu}, we introduce a likelihood-based Yes/No QA approach for zero-shot fine-grained image classification to determine the aligned class description for a given image. Given an image \(I\), we formulate a query \(Q(D_c)\) that prompts the LVLM to generate a binary response ("Yes" or "No") based on the alignment between a candidate class description \(D_c\) and the image. We use the probability of generating a "Yes" response as a confidence score to choose the best-matched class description for that image. Formally, for each candidate class \(c \in \mathcal{C}\), we compute:
\begin{equation}
    P_{\text{Yes}}(c) = p(\text{"Yes"} \mid I, Q(D_c))
\end{equation}
\noindent
where \(P_{\text{Yes}}(c)\) represents the probability of generating "Yes" for class \(c\), indicating the model's confidence that \(D_c\) correctly describes the image.

The final predicted class \(\hat{y}\) is determined by selecting the class with the highest probability of generating "Yes":
\begin{equation}
    \hat{y} = \arg\max_{c \in \mathcal{C}} P_{\text{Yes}}(c)
\end{equation}

While this approach effectively selects class descriptions, it requires access to output token probabilities to compute the likelihood of generating "Yes", limiting its implementation to LVLMs that provide token-level log probabilities. Additionally, comparing each image against every class description is computationally expensive. To address these limitations, we propose an iterative MCQA approach that both reduces computational cost and eliminates the need for model logits access.  


\subsection{Iterative MCQA} \label{mcq}

We propose an \textit{iterative MCQA} framework for zero-shot fine-grained image classification using LVLMs. Given an image $I$ and a set of $N$ possible class descriptions, the LVLM iteratively selects a small subset of $m$ options (e.g., $m = 5$) to progressively refine its prediction.

At test time, the model operates without access to the ground-truth label. In the first iteration, it is presented with a random subset of $m$ class descriptions and selects the best-matching option. In subsequent iterations, the model retains the previously chosen option and compares it against $m{-}1$ new class descriptions that have not yet been evaluated. This process continues until all class descriptions are considered. The final selection in the last iteration is taken as the model’s prediction. For example, in a 20-class task with $m = 5$, the model is first shown the set $\{A, B, C, D, E\}$. Suppose it selects $A$; the next round compares $A$ with $\{F, G, H, I\}$. If it then chooses $G$, the subsequent round presents $\{G, J, K, L, M\}$, and so on. The model’s prediction is derived from its final selected option after iterating through the full set of classes.

While our iterative MCQA approach is entirely label-free, we incorporate an \textit{early stopping mechanism} during evaluation for computational efficiency. Specifically, if the ground-truth class appears in a subset and the model fails to select it, we terminate the process early, as the final outcome will be incorrect regardless of subsequent iterations. This early termination serves purely as an evaluation shortcut and does not influence the model's decision-making process. Algorithm~\ref{alg:zero_shot_mcq} (Appendix~\ref{mcq_ap}) formalizes the complete iterative MCQA procedure. 



\subsection{Attention Intervention}

Recent studies have demonstrated that LVLMs exhibit a bias toward language tokens over image tokens during response generation~\cite{pai, deCo, zhang2024redundancy}. While existing methods address this through token logit manipulation~\cite{damro, sid} or decoder attention intervention~\cite{pai, causalmm}, they often rely on visual contrastive decoding techniques~\cite{vcd} that require more computation. These approaches, while effective for general tasks, prove computationally inefficient for fine-grained classification using LVLMs where multiple class descriptions must be evaluated against each image.

Therefore, we propose a simple yet effective attention intervention technique that enhances visual information flow while maintaining computational efficiency. Our approach is motivated by two key findings: (1) LVLMs process critical visual information in intermediate early layers~\cite{zhang2024redundancy}, and (2) intermediate later/deep layers demonstrate better object-attribute grounding~\cite{deCo, jiang2024interpreting}. Given an LVLM with $L$ layers, we compute the average attention weights from layers 3 to $k$ and incorporate them into layers $k+1$ to $L-2$, excluding the last two layers to optimize visual information flow. Additionally, we average the attention weights from layers $k+1$ to $L-2$ and propagate this grounding information to the final two layers: 
\begin{equation}
\bar{A}_{\text{early}} = \frac{1}{k-2}\sum_{i=3}^{k} A_i
\end{equation}
\begin{equation}
A'_j = A_j + \lambda\bar{A}_{\text{early}}, \quad j \in [k+1, L-2]
\end{equation}
\noindent
where $A_i$ represents the attention weights in layer $i$, $\bar{A}_{\text{early}}$ is the mean attention pattern from selected early layers, $A'_j$ is the modified attention weights for layer $j$, and $\lambda$ is a scaling factor controlling the influence of early visual information. With this approach, we attempt to ensure enhanced visual information flow and grounded object information while maintaining computational efficiency.


\subsection{All-at-Once Approach}

Recent advances in LVLM architectures have enabled processing of significantly longer contexts, with models like Qwen2.5-VL~\cite{Qwen2.5-VL}, InternVL2.5~\cite{internvl2_5}, and GPT-4o~\cite{gpt4o} supporting extended text prompts that can accommodate hundreds or thousands of tokens. This capability presents an alternative approach to our iterative MCQA framework by allowing all class descriptions to be presented simultaneously in a single forward pass.

In this approach, we provide the complete set of $N$ class descriptions along with the input image and prompt the model to select the most appropriate description directly. The model receives a comprehensive view of all possible classes at once, potentially enabling more sophisticated reasoning about inter-class relationships and discriminative features. This method eliminates the iterative nature of our MCQA approach while maintaining the core principle of leveraging detailed class descriptions rather than generating class names.

We formulate this as a single-pass selection task where, given an image $I$ and the complete set of $N$ class descriptions $\{D_1, D_2, ..., D_N\}$, the model is prompted to identify the index of the best-matching class description:
\begin{equation}
\hat{y} = \text{LVLM}(I, \{D_1, ..., D_N\}, Q_{all})
\end{equation}
where $Q_{all}$ represents a prompt that instructs the model to select the best matching class description from all provided options simultaneously.


\section{Experiments and Evaluation}
To evaluate our proposed method, we conduct comprehensive experiments across five fine-grained image classification benchmark datasets using different LVLMs and our curated class descriptions. This section presents the details of our class description curation process, experimental setup, evaluation results, and ablation studies.

\subsection{Datasets}

For our experiments, we utilize several established fine-grained image classification datasets: CUB~\cite{cub} (200 bird species), Stanford Cars~\cite{cars} (196 car models), FGVC Aircraft~\cite{aircrafts} (102 aircraft variants), Food-101~\cite{food} (101 food categories), and Stanford Dogs~\cite{dogs} (120 dog breeds). These datasets present diverse challenges in fine-grained visual recognition across different domains including birds, dogs, cars, aircrafts, and foods. Following~\citet{finer, grain}, we conduct our experiments on randomly selected subsets of these datasets to maintain consistency with prior work, ensuring we exclude any images used in our class description curation process. For example, for the CUB dataset, we randomly select 5 images per class, resulting in 1,000 test images across 200 classes.

\begin{figure}[t]
    \centering
    \includegraphics[width=1.0\linewidth]{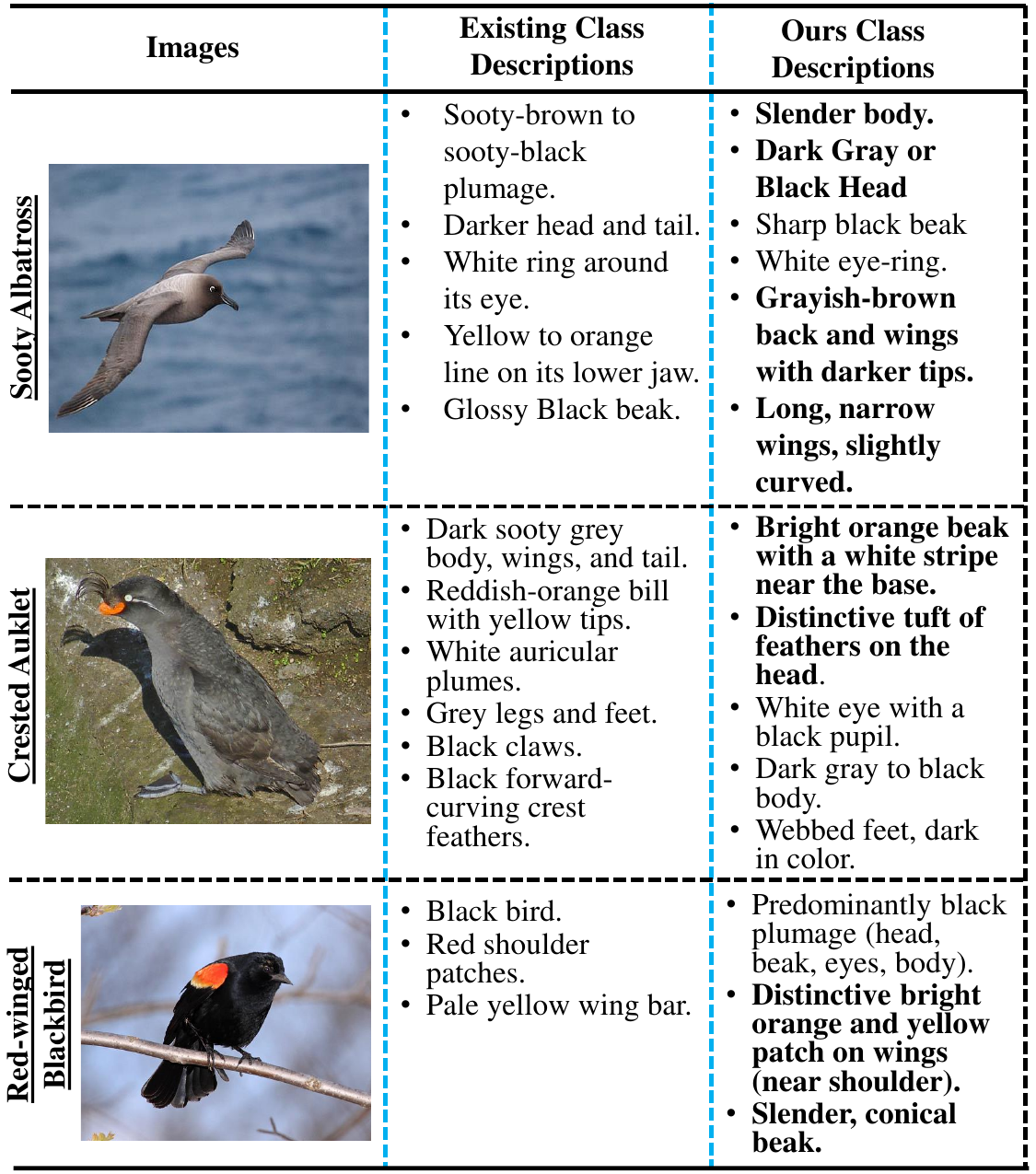}
    \caption{Comparison of class descriptions from an existing dataset and our introduced class descriptions. Bold text in the "Ours Class Descriptions" column highlights key discriminative features that are either absent or described with less specificity in the "Existing Class Descriptions" column. The increased detail in our proposed descriptions facilitates more accurate zero-shot fine-grained image classification.}
    \label{fig:data_sample}
    \vspace{-0.7cm}
\end{figure}


\subsection{Class Description Curation}

Existing class descriptions for these five datasets, provided by~\citet{menon2022visual},~\citet{saha2024improved}, and~\citet{finer}, are either generated using LLMs like GPT-3/4~\cite{menon2022visual, saha2024improved} or extracted from the internet and refined using GPT-4V~\cite{finer}. We observe that these descriptions often lack precise fine-grained visual attributes necessary for accurate identification and may contain irrelevant information.

To address these limitations, we develop a two-stage process for curating more accurate class descriptions for these five datasets (prompts in Appendix~\ref{dataset_prompts}). First, we iteratively feed individual representative images from each class (10 images total) to Qwen2.5-VL-72B-Instruct~\cite{Qwen2.5-VL}, a state-of-the-art LVLM, to generate detailed visual descriptions for each image. Then, we process these 10 descriptions through Qwen2.5-72B-Instruct~\cite{qwen2.5} LLM to compile a single comprehensive class description that captures common, distinctive visual features.

Our curation process yields 10 benchmarks - paired versions of 5 datasets with and without class names in their descriptions. Figure~\ref{fig:data_sample} demonstrates the enhanced descriptive precision of our curated descriptions for the CUB dataset~\cite{cub} compared to existing class descriptions from Finer~\cite{finer}. Our descriptions incorporate more specific visual attributes (highlighted in bold) that are essential for fine-grained classification. For example, in describing the Crested Auklet, our description specifies a "bright orange beak with a white stripe near the base" and a "distinctive tuft of feathers", providing more precise detail than the existing description's "reddish-orange bill with yellow tips" and "black forward-curving crest feathers". Similarly, for the Red-winged Blackbird, we include the "distinctive bright orange and yellow patch on wings (near shoulder)", offering crucial color and spatial information missing from the existing description's simple "red shoulder patches". This enhanced specificity across multiple visual features (beak morphology, plumage patterns, body structure) facilitates more accurate zero-shot classification among visually similar species.

\begin{table*}[!ht]
    \small 
    \centering
    \begin{tabular}{lllccccc}   \toprule
     \textbf{Setting} & \textbf{Method} & \textbf{Model} & \textbf{CUB} & \textbf{\makecell{Stanford\\Dogs}} & \textbf{\makecell{Stanford\\Cars}} & \textbf{\makecell{FGVC\\Aircraft} }&\textbf{\makecell{Food\\101}} \\ \midrule
     
     \multirow{20}{*}{W/o name } &\multirow{4}{*}{Yes/No QA}  &LLaVA-v1.5-7B &11.10 &9.50 &14.18 &7.43 &52.28  \\
     & &LLaVA-v1.5-13B &13.80  &10.50 &18.67 &9.71 &56.24 \\ 
     & &Qwen2-VL-7B &20.20 &22.33 &21.94 &13.71 &67.33 \\ 
     & &Qwen2.5-VL-7B &21.30 &18.00 &22.04 &10.57 &65.15 \\ \cmidrule(lr){2-8}

     & \multirow{11}{*}{Iterative MCQA} &LLaVA-v1.5-7B & 9.00 & 15.67 & 9.90 & 11.71 &40.99  \\
      & &LLaVA-v1.5-7B + Attn & 12.10 & 10.17 & 14.29 & 15.14  & 43.76 \\
      & & \cellcolor{lightgray} $\Delta$ & \cellcolor{lightgray}+3.10 
        & \cellcolor{lightgray}-5.50 & \cellcolor{lightgray}+4.39 
        & \cellcolor{lightgray}+3.43 & \cellcolor{lightgray}+2.77 \\
      & &LLaVA-v1.5-13B & 7.50 & 16.00 & 13.88 & 10.57 & 32.87 \\
      & &LLaVA-v1.5-13B + Attn & 8.80 & 16.33 & 15.10 & 12.29 & 30.30 \\
      & & \cellcolor{lightgray} $\Delta$ & \cellcolor{lightgray}+1.30 
        & \cellcolor{lightgray}+0.33 & \cellcolor{lightgray}+1.22 
        & \cellcolor{lightgray}+1.72 & \cellcolor{lightgray}-2.57 \\
      & &InternVL2\_5-8B & 26.50 & 21.00 & 27.35 & 12.86 & 58.42\\
      & &InternVL2\_5-8B + Attn & 27.30 & 21.67 & 27.65 & 13.71 &58.81 \\
      & & \cellcolor{lightgray} $\Delta$ & \cellcolor{lightgray}+0.80 
        & \cellcolor{lightgray}+0.67 & \cellcolor{lightgray}+0.30 
        & \cellcolor{lightgray}+0.85 & \cellcolor{lightgray}+0.39 \\
      & & Qwen2-VL-7B & 32.40 & 15.67 &33.78 & 20.00 & 54.85 \\ 
      & &Qwen2.5-VL-7B & 30.60 & 21.50 & 41.63 & 17.71 & 63.76 \\  \cmidrule(lr){2-8}

      & \multirow{5}{*}{All-at-Once} &InternVL2\_5-8B & 15.30 & 17.67 & 16.22 & 10.57 & 55.84 \\
      & & Qwen2.5-VL-7B & 6.60 & 14.33 & 21.94 & 12.00 & 43.76 \\  
      & & Qwen2.5-VL-32B & 15.60 & 23.83 & 47.96 & 21.43 & 49.31 \\ 
      
      & & GPT-4o &35.00 & 50.00 &50.51 &  27.14 & 80.20  \\ 
      & & o4-mini (reasoning) &\textbf{ 48.50} & \textbf{58.33} & \textbf{64.29} &\textbf{40.00} & \textbf{89.11} \\ \midrule

      
     \multirow{20}{*}{With name} &\multirow{4}{*}{Yes/No QA} &LLaVA-v1.5-7B &15.60 &17.00  &36.22  &19.14 &55.84  \\  
     & &LLaVA-v1.5-13B &15.60 &22.17 &36.02 &23.14 &61.98\\ 
     & &Qwen2-VL-7B &53.30 & 62.33 &\textbf{86.84} &63.43 &75.45 \\  
     & &Qwen2.5-VL-7B &53.80 &58.00 &75.71 &49.14 &67.92 \\ \cmidrule(lr){2-8}

     & \multirow{11}{*}{Iterative MCQA} &LLaVA-v1.5-7B & 18.40 & 19.83 & 20.20 & 30.86 & 47.92 \\
      & &LLaVA-v1.5-7B + Attn & 23.30 & 22.50 & 21.94 & 31.14  & 48.32 \\
      & & \cellcolor{lightgray} $\Delta$ & \cellcolor{lightgray}+4.90
        & \cellcolor{lightgray}+2.67 & \cellcolor{lightgray}+1.74
        & \cellcolor{lightgray}+0.28 & \cellcolor{lightgray}+0.40 \\
      & &LLaVA-v1.5-13B & 12.60 & 29.83 & 32.45 & 18.00 & 40.59 \\
      & &LLaVA-v1.5-13B + Attn & 13.80 & 31.33 & 31.43 & 18.57 & 37.43 \\
      & & \cellcolor{lightgray} $\Delta$ & \cellcolor{lightgray}+1.20 
        & \cellcolor{lightgray}+1.50 & \cellcolor{lightgray}-1.02 
        & \cellcolor{lightgray}+0.57 & \cellcolor{lightgray}-3.16 \\
      & &InternVL2\_5-8B & 36.60 & 38.50 & 42.65 & 33.43 & 56.63 \\
      & &InternVL2\_5-8B + Attn & 37.50 & 38.00 & 42.86 & 32.57 & 57.03 \\
      & & \cellcolor{lightgray} $\Delta$ & \cellcolor{lightgray}+0.90 
        & \cellcolor{lightgray}-0.50 & \cellcolor{lightgray}+0.21 
        & \cellcolor{lightgray}-0.86 & \cellcolor{lightgray}+0.40 \\
      & & Qwen2-VL-7B & 53.20 & 46.33 &71.12 & 54.57 & 54.26 \\ 
      & &Qwen2.5-VL-7B & 53.00 & 46.17 & 65.92 & 52.86 & 67.13  \\  \cmidrule(lr){2-8}

    & \multirow{5}{*}{All-at-Once} &InternVL2\_5-8B & 20.60 & 30.00 & 21.84 & 24.86 & 57.82 \\
    & & Qwen2.5-VL-7B & 15.20 & 31.33 & 31.63 & 40.00 & 51.49\\
    & & Qwen2.5-VL-32B & 33.00 & 52.67 & 62.24 & 42.86 & 55.05 \\     
    & & GPT-4o & \textbf{70.50} & \textbf{74.17} & 80.10 & \textbf{77.14} & 90.10 \\ 
    & & o4-mini (reasoning) & 67.20 & 71.67 & 83.16 & \textbf{77.14} & \textbf{91.09} \\ \bottomrule
     
    \end{tabular}
    \caption{Zero-shot fine-grained visual classification performance in accuracy using different methods and models. Results compare Yes/No QA, iterative MCQA, and all-at-once approaches, with our proposed attention intervention (+Attn) enhancement. We conducted experiments with and without class names in descriptions on five fine-grained classification benchmarks. $\Delta$ rows show the performance gains from our introduced attention intervention. Best performance in each setting is highlighted in \textbf{bold}.}
    \label{tab:main_results}
    \vspace{-0.3cm}
\end{table*}


\subsection{Experimental Setup}

For our experiments, we employ five state-of-the-art LVLMs: LLaVA-v1.5-7B, LLaVA-v1.5-13B, InternVL2\_5-8B, Qwen2-VL-7B-Instruct (also referred to as Qwen2-VL-7B in this literature), and Qwen2.5-VL-7B-Instruct. For our attention intervention experiments, we set $\lambda$=1 and $k$=21 for LLaVA-v1.5-7B, InternVL2\_5-8B models and $k$=23 for LLaVA-v1.5-13B model. We evaluate model performance using classification accuracy as our primary metric.


\subsection{Results}
We evaluate our proposed approaches: Yes/No QA, iterative MCQA, iterative MCQA with attention intervention, and all-at-once across multiple LVLMs on multiple fine-grained classification benchmarks. Table~\ref{tab:main_results} presents comprehensive results under two experimental settings: with and without class names in class descriptions.

\textbf{Comparative analysis of proposed methods.}
Our iterative MCQA approach consistently outperforms the Yes/No QA baseline across all models and experimental settings. In the "without name" setting, Qwen2.5-VL-7B-Instruct achieves 30.60\% on CUB with iterative MCQA compared to 21.30\% with Yes/No QA, representing a substantial 9.30\% improvement. We observe a similar trend for Stanford Cars (41.63\% vs 22.04\%) and FGVC Aircraft (17.71\% vs 10.57\%) datasets with the Qwen2.5-VL-7B-Instruct model. The inclusion of class names dramatically shifts the performance landscape. Yes/No QA becomes highly competitive, with Qwen2-VL-7B-Instruct achieving exceptional results (53.30\% on CUB, 86.84\% on Stanford Cars, the latter being the best performance across all methods). Meanwhile, iterative MCQA maintains strong performance while offering the advantage of structured multi-step reasoning, particularly evident in models like InternVL2\_5-8B achieving balanced performance across all datasets.

\textbf{Impact of Attention Intervention.}
The attention intervention shows variable effectiveness, with improvements varying across architectures and datasets. LLaVA-v1.5-7B demonstrates the most consistent gains from attention intervention, particularly in the "with name" setting where it achieves +4.90\% improvement on CUB (18.40\% → 23.30\%). LLaVA-v1.5-13B shows moderate improvements ranging from +1.20\% to +1.72\% across most datasets, though with some minor decreases on Stanford Cars and Food-101.
InternVL2\_5-8B exhibits the most stable response to attention intervention, with modest but consistent gains (+0.21\% to +0.90\%) across datasets in both settings. We performed layer selection optimization for the LLaVA-v1.5-7B model using the CUB dataset and applied a similar layer selection to the larger models. As we optimized layer selection specifically for LLaVA-v1.5-7B, the intervention proves most beneficial for this model. However, the fact that similar layer selection also shows performance gains for other models indicates that model-specific and dataset-specific layer selection could further enhance the effectiveness of our attention intervention method.

\textbf{All-at-once vs. iterative approach.}
Our all-at-once approach reveals interesting trade-offs between context utilization and performance. While state-of-the-art proprietary models (GPT-4o: 70.50\% on CUB, o4-mini: 67.20\% on CUB) achieve competitive results, open-source models suffer significant performance degradation when processing all class descriptions simultaneously. The stark contrast is evident in Qwen2.5-VL-7B-Instruct, which achieves 53.00\% on CUB with iterative MCQA but only 15.20\% with all-at-once in the "with name" setting, a 37.80\% performance drop that highlights the cognitive burden of simultaneous multi-class comparison.

\textbf{Key Findings.}
Four critical insights emerge from our evaluation: (1) Iterative MCQA consistently outperforms Yes/No QA, with average improvements ranging from 5\% to 20\% across datasets; (2) Attention intervention effectiveness varies across different model architectures, with LLaVA-v1.5-7B benefiting more than InternVL2\_5-8B, particularly when using our optimized layer selection tailored for the LLaVA-v1.5-7B model on the CUB dataset; (3) Including class names significantly enhances performance across all approaches, with particularly notable improvements on FGVC Aircraft and Stanford Cars datasets; (4) The all-at-once approach works well for proprietary models but shows substantial degradation in open-source models, highlighting the advantage of iterative processing. These results demonstrate that structured, iterative processing with attention guidance provides superior performance for zero-shot fine-grained classification compared to single-pass approaches.


\subsection{Comparison} 

Table~\ref{tab:performance_comp} presents a performance comparison between our proposed method and existing approaches across four fine-grained visual classification datasets. Our method significantly outperforms Finer~\cite{finer}, the current SOTA approach, across all datasets. While Finer achieves accuracies of 1.56\%, 4.20\%, 0.50\%, and 1.50\% on CUB, Stanford Dogs, Stanford Cars, and FGVC Aircraft, respectively, our approaches demonstrate substantial improvements across all datasets.

Our iterative MCQA approach with attention intervention (MCQA + Attn) achieves SOTA performance on most datasets: CUB (23.30\%), Stanford Dogs (22.50\%) and Aircraft (31.14\%), with absolute accuracy gains of 21.74\%, 18.30\%, and 29.64\% over Finer, respectively. Meanwhile, the Yes/No QA approach excels on Stanford Cars with the highest accuracy of 36.22\%, achieving a substantial 35.72\% improvement over the baseline.

These significant improvements, with absolute accuracy gains ranging from 18.30\% to 35.72\% over Finer, validate that our proposed approaches effectively enhance the model's ability to focus on discriminative visual features for fine-grained classification. Among our proposed approaches, the iterative MCQA with attention intervention demonstrates the most consistent performance across multiple datasets. A detailed comparison with CLIP-based methods is provided in Appendix~\ref{clip_compare}, and a discussion of the computational efficiency of our approach relative to the baselines is included in Appendix~\ref{algo_efficiency}.

\begin{table}[!ht]
    \centering
    \fontsize{8}{10}\selectfont 
    \setlength{\tabcolsep}{2.5pt} 
    \begin{tabular}{lcccc}     \toprule
     \textbf{Method}   & \textbf{CUB} & \textbf{\makecell{Stanford\\Dogs}} & \textbf{\makecell{Stanford\\Cars}}   &\textbf{\makecell{FGVC\\Aircraft}} \\ \midrule
     Finer~\cite{finer}  &1.56  &4.20 &0.50 &1.50\\
     Ours (Yes/No QA) &15.60 &17.00 &\textbf{36.22} &19.14 \\
     Ours (MCQA) &18.40 &19.83 &20.20  &30.86 \\
     Ours (MCQA + Attn) &\textbf{23.30} &\textbf{22.50 } &21.94 &\textbf{31.14} \\ \bottomrule
    \end{tabular}
   \caption{Performance comparison of our proposed method with existing methods. For a fair comparison, all experiments use the LLaVA-v1.5-7B model. For Finer, values are extracted from the published results graphs with a potential variance of 1-2\%. Best performance for each dataset is highlighted in \textbf{bold}.}
    \label{tab:performance_comp}
    \vspace{-0.5cm}
\end{table}


\subsection{Ablation Study}
We conduct a series of ablation studies to evaluate the impact of key components in our proposed method. Additional ablation results are provided in Appendix~\ref{options_mcq}, and~\ref{Im_prompt}.

\subsubsection{Comparison of Datasets}
Table~\ref{tab:dataset_comparison} highlights the importance of our introduced class descriptions in zero-shot fine-grained classification using Qwen2.5-VL-7B-Instruct~\cite{Qwen2.5-VL} in the iterative MCQA setting. The existing class descriptions, sourced from the Finer~\cite{finer} (except for Food-101, which comes from~\citet{menon2022visual}), lack detailed attribute representation, leading to lower accuracy. In contrast, our introduced descriptions, which emphasize precise visual attributes, significantly improve performance, with notable gains on Food-101 (+13.07\%) and CUB (+8.3\%). This demonstrates that richer attribute-based descriptions enhance the model’s ability to distinguish fine-grained classes.  

Including class names further improves accuracy across all datasets, as the model benefits from explicit textual cues. However, even in this setting, our descriptions consistently outperform the existing ones by up to +30.30\%, demonstrating the effectiveness of structured visual attributes. Interestingly, the performance gap narrows with class names, suggesting that while names aid classification, detailed descriptions remain crucial for capturing fine-grained distinctions. This underscores the importance of precise textual descriptions for improving zero-shot classification in LVLMs.

\begin{table}[!ht]
    \centering
    \fontsize{8}{10}\selectfont 
    \setlength{\tabcolsep}{2.5pt} 
    \begin{tabular}{lccccccc}
        \toprule 
        \multirow{2}{*}{\textbf{Dataset}} & \multicolumn{3}{c}{\textbf{Without Class Name}} & \multicolumn{3}{c}{\textbf{With Class Name}} \\
        \cmidrule(lr){2-4} \cmidrule(lr){5-7} \cmidrule(lr){6-7}
        & Existing & Ours &$\Delta$ & Existing & Ours &$\Delta$ \\
        \midrule
        CUB           & 22.30 & 30.60 & +8.30 & 50.10 & 53.00  & +2.90 \\
        Stanford Dogs    & 14.50 & 21.50 & +7.00 & 44.00 & 46.17  & +2.17 \\
        Stanford Cars    & 42.55 & 41.63 &-0.92 & 65.20 & 65.92  & +0.72 \\
        FGVC Aircraft   & 14.86 & 17.71 & +2.85 & 51.71 & 52.86  & +1.15 \\
        Food-101         & 50.69 & 63.76 & +13.07 & 36.83 & 67.13  & +30.3 \\
        \bottomrule 
    \end{tabular}
    \caption{Performance comparison of Qwen2.5-VL-7B-Instruct model with iterative MCQA approach on existing class descriptions and our introduced class descriptions, with and without the class name in the class descriptions. The $\Delta$ columns show the performance gain of our introduced class descriptions over the existing ones in both settings.}
    \label{tab:dataset_comparison}
    \vspace{-0.5cm}
\end{table}

\subsection{Impact of Model Scale} \label{impact_model_scale}
To evaluate the effect of model capacity, we conducted experiments using Qwen2.5-VL-7B-Instruct and Qwen2.5-VL-32B-Instruct in our all-at-once approach. As shown in Table~\ref{tab:main_results}, performance consistently improves with increased model size across most datasets and settings. In the "with name" setting, the 32B model demonstrates substantial improvements over the 7B variant: CUB (33.00\% vs 15.20\%), Stanford Dogs (52.67\% vs 31.33\%), Stanford Cars (62.24\% vs 31.63\%), FGVC Aircraft (42.86\% vs 40.00\%), and Food-101 (55.05\% vs 51.49\%). In the "without name" setting, similar trends are observed with significant improvements on Stanford Dogs (23.83\% vs 14.33\%) and Stanford Cars (47.96\% vs 21.94\%), while FGVC Aircraft shows notable enhancement (21.43\% vs 12.00\%). The results highlight that larger models generally provide better capacity for distinguishing fine-grained visual features, particularly when class names are available to provide additional semantic context. This scaling benefit is most pronounced on datasets requiring detailed visual discrimination, such as Stanford Cars and Stanford Dogs.


\subsubsection{Robustness of Iterative MCQA}
To assess the robustness of our iterative MCQA approach, we analyze how often models deviate from providing the requested index numbers by giving non-numerical responses. Our experiments show high instruction adherence across models. Among many experiments, LLaVA models produced only 2-43 non-numerical responses across 8 experiments, each involving 3,000-20,000 generation calls. Similarly, InternVL2\_5-8B and Qwen2.5-VL models showed strong reliability with just 3-19 such errors in 3 experiments. These minimal error rates are negligible compared to the total number of model calls, and the rare instances of non-compliance could be addressed by directly computing probabilities for option indices. In contrast, the all-at-once approach showed more formatting issues, with Qwen2.5-VL and InternVL2\_5-8B models occasionally providing non-numerical values or responses with explanatory text. The consistently low error rates in iterative MCQA validate that our approach is both efficient and robust.


\section{Conclusion}

The application of Large Vision-Language Models (LVLMs) to zero-shot fine-grained image classification represents a promising direction in vision-language understanding. We address current limitations by proposing multiple approaches: Yes/No QA, iterative multiple-choice question answering (MCQA), and all-at-once methods, enhanced by a simple yet effective attention intervention technique. Our investigation revealed the limitations of existing LLM-generated class descriptions, prompting us to curate refined descriptions that better capture fine-grained visual characteristics. Through comprehensive experiments across five datasets and multiple model architectures, we demonstrate that our iterative MCQA method consistently outperforms the Yes/No QA approach, while the all-at-once approach shows promising results for proprietary models but limitations for open-source architectures. Our methods achieve substantial performance improvements, significantly outperforming existing SOTA approaches across all evaluated datasets. The effectiveness of our attention intervention mechanism, particularly its benefits for smaller models, suggests promising directions in developing more sophisticated attention guidance techniques tailored to specific LVLM architectures for zero-shot fine-grained classification tasks.


\section{Limitations}
While our iterative MCQA approach with attention intervention demonstrates promising results in zero-shot fine-grained visual classification, several key limitations exist. The lower performance on most datasets indicates that current LVLMs still struggle with fine-grained visual distinctions. Additionally, our attention intervention technique's effectiveness varies across architectures and datasets, suggesting dependency on model architecture and task characteristics. The quality of class descriptions is crucial, and despite leveraging advanced LVLMs and LLMs, they may still miss discriminative features that domain experts would identify. Refining these descriptions requires substantial expertise and manual effort, limiting scalability. Like other zero-shot approaches, our method's effectiveness is bounded by the pre-training data quality and coverage of LVLMs, particularly for specialized domains underrepresented in the training data.

\section{Acknowledgments}
This project was supported in part by the Virginia Tech College of Science Academy of Data Science Discovery Fund (ADSDF) (Award: 238695) and the Research Acceleration Program (RAP) grant at Carilion Clinic, Roanoke, Virginia, USA. We also acknowledge Advanced Research Computing at Virginia Tech for providing computational resources and technical support that have contributed to the results reported within this paper. We thank all reviewers for their comments, which helped improve the paper.


\appendix

\section{Appendix}
\label{sec:appendix}
This section contains the details of the following topics:
\begin{itemize}[noitemsep]
    \item Prompt for iterative MCQA (Appendix~\ref{prompt_mcq})
    \item Prompts for our Class Descriptions Generation (Appendix~\ref{dataset_prompts})
    \item Algorithm of our Iterative MCQA (Appendix~\ref{mcq_ap}
    \item Impact of Number of Options in Iterative MCQA Approach (Appendix~\ref{options_mcq})
    \item Impact of Prompts (Appendix~\ref{Im_prompt})
    \item Efficiency of the Iterative MCQA Framework (Appendix~\ref{algo_efficiency})
    \item Comparison with CLIP-based Methods (Appendix~\ref{clip_compare})
\end{itemize}


\subsection{Prompt for iterative MCQA} \label{prompt_mcq}

\begin{figure}[!ht]
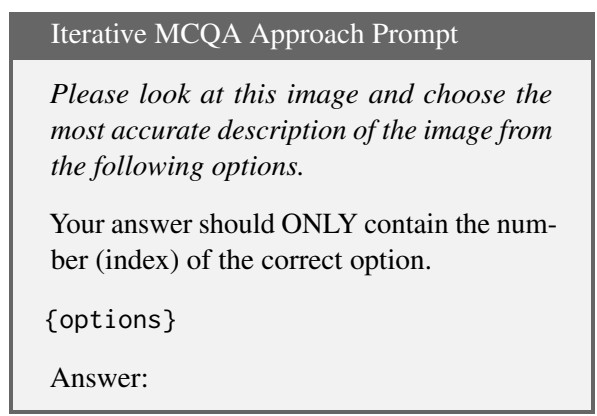

    \centering
    \begin{tcolorbox}[colback=gray!10, colframe=gray!80!black, width=\linewidth, sharp corners, title=Iterative MCQA Approach Prompt]
        \textit{Please look at this image and choose the most accurate description of the image from the following options.}

        \vspace{0.3cm}
        Your answer should ONLY contain the number (index) of the correct option.

        \vspace{0.3cm}
        \texttt{{\{options\}}}

        \vspace{0.3cm}
        Answer:
    \end{tcolorbox}
    \caption{Prompt used for the iterative multiple-choice question answering (MCQA) approach.}
    \label{fig:mcq-prompt}
\end{figure}

For our proposed method, we use a consistent prompt template across all experiments. Figure~\ref{fig:mcq-prompt} shows the prompt used in our iterative MCQA approach for zero-shot fine-grained image classification.

\noindent
Our prompt structure enforces methodological consistency across datasets and experimental settings while constraining model outputs to numerical indices.


\subsection{Prompts for our Class Descriptions Generation} \label{dataset_prompts}

\begin{figure}[!ht]
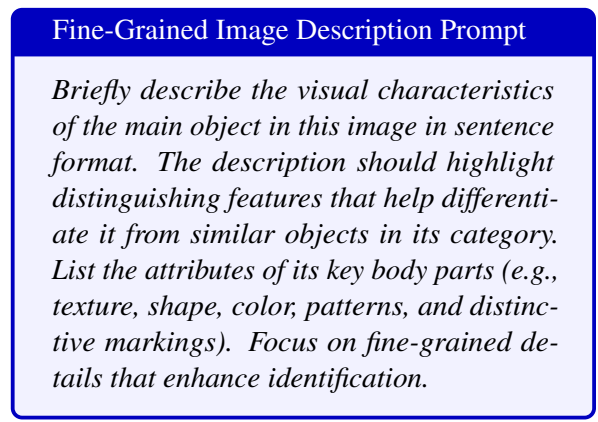

    \centering
    \begin{tcolorbox}[colback=blue!5!white, colframe=blue!75!black, title=Fine-Grained Image Description Prompt]
        \textit{Briefly describe the visual characteristics of the main object in this image in sentence format. The description should highlight distinguishing features that help differentiate it from similar objects in its category. List the attributes of its key body parts (e.g., texture, shape, color, patterns, and distinctive markings). Focus on fine-grained details that enhance identification.}
    \end{tcolorbox}
    \caption{Prompt used for obtaining fine-grained visual descriptions from an LVLM.}
    \label{fig:prompt-box}
\end{figure}

We developed a two-stage process for curating more accurate class descriptions:

First, we feed 10 representative images from each class to Qwen2.5-VL-72B-Instruct~\cite{Qwen2.5-VL}, a state-of-the-art LVLM, to generate detailed visual descriptions for each image. Figure~\ref{fig:prompt-box} shows used prompt for this step. 

\begin{figure}[!ht]
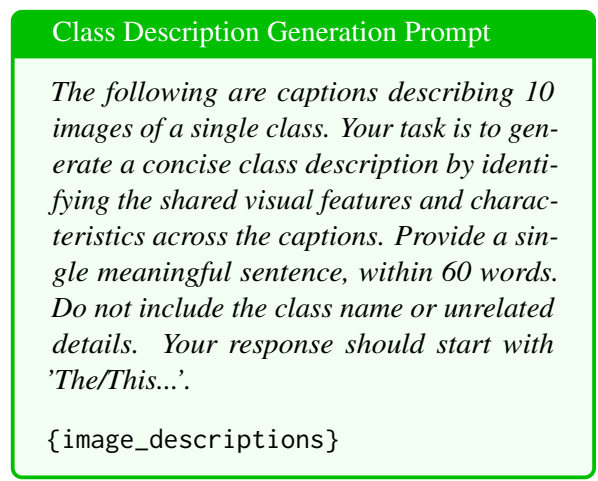

    \centering
    \begin{tcolorbox}[colback=green!5!white, colframe=green!75!black, title=Class Description Generation Prompt]
        \textit{The following are captions describing 10 images of a single class. Your task is to generate a concise class description by identifying the shared visual features and characteristics across the captions. Provide a single meaningful sentence, within 60 words. Do not include the class name or unrelated details. Your response should start with 'The/This...'.}
        
        \vspace{0.3cm}
        \texttt{{\{image\_descriptions\}}}
    \end{tcolorbox}
    \caption{Prompt used for generating class descriptions from fine-grained image captions.}
    \label{fig:class-description-prompt}
\end{figure}

Then, we process these 10 descriptions through Qwen2.5-72B-Instruct~\cite{qwen2.5} LLM to synthesize a single comprehensive class description that captures common, distinctive visual features. Figure~\ref{fig:class-description-prompt} depicts our employed prompt for this step. In the prompt, \texttt{{\{image\_descriptions\}}} represents the 10 image descriptions we obtained from the previous step.

\subsection{Algorithm of our Iterative MCQA} \label{mcq_ap}

This section presents our introduced iterative MCQA algorithm. Algorithm~\ref{alg:zero_shot_mcq} outlines the steps used in our zero-shot fine-grained classification task with LVLMs. Given an image $I$ and a set of $N$ possible class descriptions, the method iteratively selects a subset of $m$ descriptions and queries the model to identify the best match. The process continues until either all class descriptions have been evaluated or a misclassification occurs, with the predicted class retained in the next $(m{-}1)$ descriptions. A prediction is considered correct only if the ground-truth class is consistently identified throughout all iterations (i.e., in the final prediction). Further details are provided in Section~\ref{mcq}.

\begin{algorithm}[!ht]
\caption{Iterative Multiple-Choice Question Answering for Zero-Shot Image Classification}
\label{alg:zero_shot_mcq}
\begin{algorithmic}[1]
\Require Image $I$, Set of Class Descriptions $C = \{c_1, c_2, ..., c_n\}$, Model $M$, Batch Size $m$
\Ensure Predicted Class $c^*$

\State Initialize $correct \leftarrow False$, $processed \leftarrow False$

\While{not $processed$}
\State Select $m$ candidate descriptions $C_m$ 
\State Query $M$ with $I$ and $C_m$, obtaining predicted class $c_p$ 
\If{$c \in C_m$ and $c_p \neq c$}
\State \Return Misclassification \Comment{Terminate early as $M$ missed the correct class}

\Else
\State Move to next $m$ descriptions keeping $c_p$ in $m$ descriptions.
\EndIf
\If{all class descriptions are processed}
\State $processed \leftarrow True$ 
\EndIf
\EndWhile
\If{$c_p = c$ and $processed$}
\State \Return Classified Correctly 
\Else
\State \Return Misclassified 
\EndIf
\end{algorithmic}

\end{algorithm}

\begin{table*}[!ht]
\centering
\small
\begin{tabular}{lcccc}
\toprule
\textbf{Method} & \textbf{CUB} & \textbf{Food-101} & \textbf{Aircraft} & \textbf{Cars} \\
\midrule
Ours & \textbf{70.50} & 91.09 & \textbf{77.14} & \textbf{86.84} \\
\citet{menon2022visual} & 65.26 & 93.26 & -- & -- \\
CuPL~\cite{pratt2023does} & -- & \textbf{93.33} & 36.69 & 76.49 \\
CascadeVLM~\cite{wei2024enhancing} & -- & -- & 36.80 & 85.60 \\
LaBo~\cite{yang2023language}  & 54.19 & 80.41 & 32.73 & -- \\
Yan et al.~\cite{yan2023learning} & 64.05 & 81.85 & -- & 81.85 \\
V-GLOSS~\cite{ogezi2024semantically} & -- & -- & 38.70 & -- \\
\bottomrule
\end{tabular}
\caption{Performance comparison (Top-1 Accuracy \%) of our method and prior CLIP-based methods across five fine-grained classification datasets. Missing values (--) indicate that the method did not report results for the corresponding dataset.}
\label{tab:clip_comparison}
\end{table*}


\subsection{Impact of Number of Options in Iterative MCQA Approach} \label{options_mcq}

\begin{table}[!ht]
    \centering
    \fontsize{9}{10}\selectfont 
    \setlength{\tabcolsep}{2.5pt} 
    \begin{tabular}{cccccc}     \toprule
     \textbf{\#Options}   & \textbf{CUB} & \textbf{\makecell{Stanford\\Cars}}  & \textbf{\makecell{Stanford\\Dogs}} &\textbf{\makecell{FGVC\\Aircraft}} &\textbf{\makecell{Food\\101}} \\ \midrule
        5 & 53.00 &46.17 &65.92 &52.82 &67.13 \\
        10 &55.50  &67.65 &43.50 &48.86 & 59.01 \\ \bottomrule

     \end{tabular}
    \caption{Impact of the number of multiple-choice options on model accuracy across different fine-grained visual classification datasets. All experiments were conducted using the Qwen2.5-VL-7B-Instruct model with class descriptions including class names.}
    \label{tab:options_impact}
\end{table}

Traditional Multiple-Choice Question Answering (MCQA) approaches typically employ 4 or 5 options. Table~\ref{tab:options_impact} presents a comparative analysis of model performance when varying the number of multiple-choice options in our iterative MCQA approach. We conducted experiments using the Qwen2.5-VL-7B-Instruct model across five fine-grained visual classification datasets.
Our choice of using 5 options was motivated by three factors: alignment with traditional MCQA practices, the performance advantages observed with this configuration, and the context length limitations of the LLaVA-v1.5 models. As shown in the results, the impact of increasing options from 5 to 10 varies significantly across datasets. While some datasets show performance degradation (Stanford Dogs: -22.42\%, FGVC Aircraft: -3.96\%, Food-101: -8.12\%), others demonstrate substantial improvements (Stanford Cars: +21.48\%, CUB: +2.50\%).

Notably, Stanford Cars shows the most dramatic improvement with 10 options (67.65\% vs 46.17\%), suggesting that for certain datasets with high inter-class similarity, additional options may help the model make more precise distinctions. Conversely, Stanford Dogs exhibits the largest performance drop (-22.42\%), indicating that too many options can overwhelm the model's decision-making process for certain classification tasks. With 5 options, the model achieved accuracies ranging from 46.17\% to 67.13\% across different datasets, while with 10 options, the range was 43.50\% to 67.65\%.

\subsection{Impact of Prompts} \label{Im_prompt}
We investigated the effect of prompt engineering on our iterative Multiple Choice Question Answering (MCQA) approach. Specifically, we enhanced the existing prompt by adding the instruction: \textit{"The options are fine-grained. You have to utilize your grounding capability or visual attribute matching capability to select the best-aligned option."} This modification, designed to encourage more precise visual analysis, yielded mixed results when tested on our Qwen2.5-VL-7B-Instruct model.

For the Stanford Dogs dataset, we observed modest improvements: accuracy increased from 46.17\% to 48.33\% with class names provided, and from 21.50\% to 22.67\% without class names, respectively. A similar improvement was observed for the FGVC Aircraft dataset with class names, where accuracy improved from 52.86\% to 54.57\%. However, performance decreased for some datasets: the Food-101 dataset accuracy declined from 67.13\% to 65.35\% with class names.

An important trade-off emerged during testing: while the modified prompt generally improved accuracy, it also affected response consistency. Without the added instruction, the model consistently complied with the format requirement of returning only the option index. However, after introducing the new instruction, we observed that in 10 to 28 additional instances beyond our normal prompt, the model deviated from this format, providing additional explanatory text along with the index. This suggested that while the instruction helped improve accuracy in most cases, it compromised the model's adherence to output formatting requirements.



\subsection{Efficiency of the Iterative MCQA Framework} \label{algo_efficiency}
Our proposed iterative multiple-choice question answering (MCQA) framework provides a favorable balance between classification accuracy and computational efficiency. While traditional Yes/No QA approaches require $\mathcal{O}(N)$ forward passes for $N$ class descriptions—one per class—our MCQA method reduces the number of passes to approximately $N/4$ by employing 5-way batching. For instance, in a dataset with 200 classes such as CUB, Yes/No QA requires 200 separate forward passes, whereas the iterative MCQA framework typically completes classification in only 40--50 passes, achieving a 75--80\% reduction in computation. As shown in Table~\ref{tab:options_impact}, increasing the batch size beyond 5 options results in decreased accuracy, indicating that 5-way MCQA provides a practical trade-off between inference cost and performance. These results demonstrate that the iterative MCQA method enables scalable zero-shot fine-grained classification with significantly lower computational overhead.

\subsection{Comparison with CLIP-based Methods.}
\label{clip_compare}

We compare our zero-shot fine-grained classification framework with several recent CLIP-based and LVLM-augmented approaches across diverse datasets. Our results are based on the All-at-Once approach using GPT-4o or o4-mini. While traditional CLIP-based methods such as CuPL~\cite{pratt2023does} and \citet{menon2022visual} report high performance on Food-101 (93.33\% and 93.26\% respectively), they rely heavily on handcrafted prompts or multiple prompt augmentations. In contrast, our approach achieves competitive performance on Food-101 (91.09\%) using a unified, training-free framework.
On more fine-grained datasets, our method substantially outperforms LaBo~\cite{yang2023language} on CUB (70.50\% vs. 54.19\%) and FGVC Aircraft (77.14\% vs. 32.73\%), while achieving competitive results on Stanford Cars (83.16\% vs. 81.85\% from~\citet{yan2023learning}). Notably, CascadeVLM~\cite{wei2024enhancing}, which combines CLIP with LVLMs, reports higher performance on Stanford Cars (85.60\%) but much lower accuracy on FGVC Aircraft (36.80\%), indicating limited generalizability. Our approach does not rely on any supervised classifier or handcrafted prompts, yet provides robust and consistent performance across multiple domains, highlighting its generalizability and effectiveness for zero-shot fine-grained image classification.


\end{document}